\documentclass[runningheads]{llncs}
\usepackage{graphicx}
\usepackage{amsmath}
\usepackage{booktabs}
\usepackage{tikz}
\def\checkmark{\tikz\fill[scale=0.4](0,.35) -- (.25,0) -- (1,.7) -- (.25,.15) -- cycle;}
\begin{document}
\title{YOLOrtho: A Unified Framework for Teeth Enumeration and Dental Disease Detection}
%
%
\author{Shenxiao Mei,
Chenglong Ma,
Feihong Shen,
Huikai Wu,
Kaidi Shen
}
\authorrunning{Mei et al.}
%
\institute{Chohotech Inc.}
%
\maketitle              
\begin{abstract}
Detecting dental diseases through panoramic X-rays images is a standard procedure for dentists. Normally, a dentist need to identify diseases and find the infected teeth. While numerous machine learning models adopting this two-step procedure have been developed, there has not been an end-to-end model that can identify teeth and their associated diseases at the same time. To fill the gap, we develop YOLOrtho, a unified framework for teeth enumeration and dental disease detection. We develop our model on Dentex Challenge 2023 data, which consists of three distinct types of annotated data. The first part is labeled with quadrant, and the second part is labeled with quadrant and enumeration and the third part is labeled with quadrant, enumeration and disease. To further improve detection, we make use of Tufts Dental public dataset. To fully utilize the data and learn both teeth detection and disease identification simultaneously, we formulate diseases as attributes attached to their corresponding teeth. Due to the nature of position relation in teeth enumeration, We replace convolution layer with CoordConv in our model to provide more position information for the model. We also adjust the model architecture and insert one more upsampling layer in FPN in favor of large object detection. Finally, we propose a post-process strategy for teeth layout that corrects teeth enumeration based on linear sum assignment. Results from experiments show that our model exceeds large Diffusion-based model. All data using in this research paper is publicly open.
\keywords{YOLO  \and Multi-Label Object Detection \and Panoramic Dental X-ray}
\end{abstract}
\section{Introduction}
The development of deep learning techniques has been significant over the past few years, and machine learning models have been depolyed in various aspects of medical image analysis like \cite{ronneberger2015u}. Meanwhile, dentists usually spend both significant amount of time and efforts in panoramic dental image analysis to establish a good understanding in their patients' dental health conditions. The diagnosis procedure can be facilitated by usage of deep learning models. Instead of starting from scratch, dentists can build their analysis upon the initial analysis generated automatically by the deep learning model. While many attempts in this area haven been made \cite{hwang2019overview}, almost all of them have done only the individual analysis work such as quadrant detection \cite{zhao2020tsasnet}, teeth instance segmentation \cite{chung2021individual} or dental disease analysis \cite{zhu2022cariesnet}. We plan to construct a unified framework that generates output including both teeth detection and their associated dental diseases. However, the cost of collecting data that has both annotation of teeth and diseases is expensive as labeling every tooth and the associated disease requires expertise and several rounds of revision. Collecting data that contains either teeth enumeration or dental disease analysis, one the other hand, requires less time and efforts. Also, there are a few existing public dataset that is annotated with such information \cite{panetta2021tufts,hamamci2023dentex}. In this paper, we use the dataset provided in Dentex Challenge 2023 \cite{hamamci2023dentex} and Tufts Dental public dataset \cite{panetta2021tufts}. Dentex dataset is constructed in three parts. The first part is labeled with quadrant, and the second part is labeled with quadrant and enumeration and the third part is labeled with quadrant, enumeration and disease. To further improve tooth detection, we make use of Tufts Dental public dataset, which is labeled the same way as the second part of dentex dataset. It is worthwhile to note that, since the data we are using do not follow the same labeling protocol, conventional single-class object detection network generally do not work well. To cope with multi-class detection, one need to train two models designed to perform individual task and map the results together in the post-process. To improve this two-step procedure, We construct a new architecture that allows us to train tooth detection and disease diagnosis simultaneously. We build our model upon YOLO framework \cite{redmon2016you} and add additional heads to predict the attributes of an object. Another way to look at this design is that, the class of an object can be considered as an attribute, and we simply add more attributes of this object such as whether this object is impacted or whether this object has caries.
\newline
\indent To train a model from partial annotations, we follow a hierarchical fashion like \cite{hamamci2023diffusion}. We compute loss base on the types of input images. If the input image has enumeration of teeth, we omit the loss of disease attributes. Since the third part of Dentex dataset only label the teeth with diseases, we first train a regular teeth object detection model to pseudo label the rest of healthy teeth and set their attributes (i.e. is\_impacted) to be false.
\newline
\indent We have submitted our proposed method to Dental Enumeration and Diagnosis on Panoramic X-rays Challenge that takes place at MICCAI 2023.

\section{Methods}
\subsection{Models Overall}
\begin{figure}[h]
    \centering
    \includegraphics[width=\textwidth]{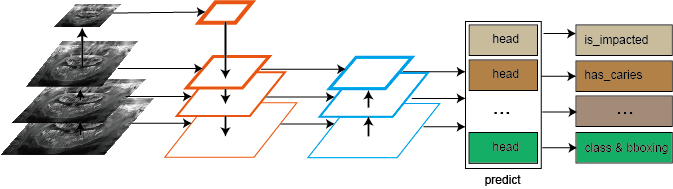}
    \caption{The architecture of our proposed framework. We apply a custom feature pyramid structure to extract the image feature and design four heads to predict desired result.}
    \label{fig:1}
\end{figure}
YOLOv8 is a well-established framework for detecting objects, and we build our model upon it. The vanilla model only supports single-class prediction for one detection. We modify the prediction heads so that multiple attributes of the detection can be predicted. In our experiment, we construct four binary classification heads: is\_impacted, has\_caries, has\_deepcaries and has\_lesion. It is worthwhile to point out that we also implement heads that provide regression and pose estimation so that this framework can be extended to a larger range of tasks. More experiments can be conducted in the future to cope with different diseases. For example, the difference of caries and deep caries is the level of structural damage, and both of these diseases share some visual features. Therefore, it is reasonable to apply regression loss on these diseases. Instead of using the default structure, we add one more upsampling in feature pyramid network. The feature map outputs of a vanilla YOLOv8 have the stride of 8, 16, 32. The close layout and uniformly large size of teeth also make them less viable to predict on feature map with a larger stride. With the modified structure, the feature map outputs have the stride of 4, 8, 16.
\subsubsection{Data Preprocess} 
We train our model to detect teeth and their associated diseases, so the training data should at least contain all teeth information. The third part of Dentex dataset only label the teeth with diseases, we first train a regular teeth object detection model to pseudo label the rest of healthy teeth and set their attributes (i.e. is\_impacted) to be false. Also, in the original annotation, if one tooth has multiple diseases, multiple bounding boxes at the same location with different disease classes are given. We combine these boxes to one box containing all information of the tooth.
\subsubsection{Data Augmentation} 
Other than the regular augmentation techniques such as scaling, random blur, rotation and translation, we implement flip mapping. If a panoramic image is flip horizontally, the quadrant information associated with the bounding boxes also changes. 

\subsection{Proposed Framework}
A conventional convolution layer work well in feature extraction and learns feature that is translation invariant. While this type of behavior is preferred in most computer vision task such as detection and classification, it can be detrimental to teeth enumeration. Position plays a huge role in teeth enumeration system. If only given relative position of a set of teeth in an image regardless of their visual cues, one can already make a near perfect enumeration prediction. To preserve positional information, we replace all convolution layer with coordinate convolution layer in the backbone network.  
\subsubsection{Loss} We add attributes heads in the framework, and these attributes prediction heads are independent of regular bounding box and classification heads. In our experiment, we have four attributes heads. For example, the prediction head for checking whether tooth is impacted looks like:$Loss_{Attribute_{is\_impacted}} = BCE(pred, GT)$. And the overall losses now becomes:
\begin{align*}
   Losses = w_b * Loss_{bbox} + w_c* Loss_{Class} + w_d * Loss_{DFL} \\ + w_1 * Loss_{Attribute_{1}}  + ... + w_n * Loss_{Attribute_{n}}  
\end{align*}
In the formula, w is a hyperparameter that stands for the weight of the corresponding loss. In the experiment we use 7.5 for bounding box, 0.5 for classification, 1.5 for DFL loss and 8 for the disease attribute loss.

\subsection{Post-Process Strategy}
To further improve the accuracy of enumeration prediction, We utilize one prior on human teeth: each FDI is associated with one tooth only. Deep learning models, on the other hand, do not have this prior knowledge. We notice that deep learning models sometimes produce same teeth enumeration for two teeth that are close to each other. To solve this problem, we formulate the enumeration post-process as a linear-sum-assignment problem: each FDI is matched with one prediction only, and the cost of objects is constructed by their probability of each class.

\section{Experiments and Results}

\begin{table}
\caption{Model Metrics in DENTEX Challenge}\label{tab1}
\centering
\begin{tabular}{|l|l|l|l|}
\hline
Model & AP-Quadrant & AP-Diagnosis & AP-Enumeration \\
\hline
Vanilla YOLO & 0.395 & 0.330 & 0.286\\
YOLOrtho & 0.414 & 0.357 & 0.337\\
HierachicalDet & 0.365 & 0.341 & 0.221\\
\hline
\end{tabular}
\end{table}

We conduct experiments on validation data in Dentex Challenge. Experiments show that YOLOrtho achieves better results in all quadrant, diagnosis and enumeration than baseline model. To the date this challenge paper is written, our method is the leading algorithm in the final test phase. Also, ablation study of YOLOrtho shows that both upsampling and the application of CoordConv significantly improve the AP metrics. The post-process strategy serves as a small trick that boosts enumeration and quadrant metrics. 

\begin{table}
  \centering
  \setlength{\tabcolsep}{12 pt}
  \resizebox{0.75\textwidth}{!}{
  \begin{tabular}{ccc|ccc}
    \toprule
    upsampling & CoordConv & post-process & AP-Quadrant & AP-Diagnosis & AP-Enumeration\\
    \midrule
    -&-&-&0.469&0.410&0.359 \\
    \checkmark&-&-&0.522&0.475&0.417 \\
    \checkmark&\checkmark&-&0.545&0.494&0.438 \\
    \checkmark&\checkmark&\checkmark&0.546 & 0.494& 0.446\\

    \bottomrule
  \end{tabular}}
  \vspace{12pt}
  \caption{Ablation study of YOLOrtho}
  \label{table:utpd}
\end{table}



%
%
%
\bibliographystyle{splncs04}
\bibliography{reference}


\end{document}